\documentclass[letterpaper]{article} 
\usepackage{main}  
\usepackage{times}  
\usepackage{helvet}  
\usepackage{courier}  
\usepackage[hyphens]{url}  
\usepackage{graphicx} 
\usepackage{natbib}  
\usepackage{caption} 
\usepackage{algorithm}
\usepackage{algorithmic}
\newcommand\Tstrut{\rule{0pt}{2.6ex}}
\usepackage{amsmath,amsfonts}
\usepackage{booktabs}
\usepackage{threeparttable}
\usepackage{multirow}
\usepackage{newfloat}
\usepackage{listings}
\DeclareCaptionStyle{ruled}{labelfont=normalfont,labelsep=colon,strut=off} 
\floatstyle{ruled}
\newfloat{listing}{tb}{lst}{}
\floatname{listing}{Listing}
\title{DifAttack: Query-Efficient Black-Box Attack via Disentangled Feature Space}
\author{
    Jun Liu\textsuperscript{1},
    Jiantao Zhou\textsuperscript{\rm 1}\thanks{Corresponding author.}
    Jiandian Zeng\textsuperscript{2},
    Jinyu Tian\textsuperscript{3}\\
}
\affiliations{
    \textsuperscript{1}
    State Key Laboratory of Internet of Things for Smart City,\\ 
    Department of Computer and Information Science, University of Macau \\

    \textsuperscript{2}
    Institute of Artificial Intelligence and Future
Networks, Beijing Normal University\\

        \textsuperscript{3}
    School of Computer Science and Engineering, Macau University of Science and Technology\\ 
   
    \{yc07453, jtzhou\}@um.edu.mo, jiandian@bnu.edu.cn, jytian@must.edu.mo

}

\begin{document}

\maketitle

\begin{abstract}

 This work investigates efficient score-based black-box adversarial attacks with a high Attack Success Rate (ASR) and good generalizability. We design a novel attack method based on a \textbf{Di}sentangled \textbf{F}eature space,  called \textbf{DifAttack}, which differs significantly from the existing ones operating over the entire feature space. Specifically, DifAttack firstly disentangles an image's latent feature into an \textit{adversarial feature} and a \textit{visual feature}, where the former dominates the adversarial capability of an image, while the latter largely determines its visual appearance. We train an autoencoder for the disentanglement by using pairs of clean images and their Adversarial Examples (AEs) generated from available surrogate models via white-box attack methods. Eventually, DifAttack iteratively optimizes the adversarial feature according to the query feedback from the victim model until a successful AE is generated, while keeping the visual feature unaltered. In addition, due to the avoidance of using surrogate models' gradient information when optimizing AEs for black-box models, our proposed DifAttack inherently possesses better attack capability in the open-set scenario, where the training dataset of the victim model is unknown. Extensive experimental results demonstrate that our method achieves significant improvements in ASR and query efficiency simultaneously, especially in the targeted attack and open-set scenarios. The code is available at https://github.com/csjunjun/DifAttack.git.

\end{abstract}

\section{Introduction}
It has been observed that meticulously crafted Adversarial Examples (AEs) are beneficial for evaluating and improving the robustness of Deep Neural Networks (DNNs) \cite{dong2020benchmarking,croce2020reliable}. AEs refer to images that deceive DNNs by incorporating imperceptible perturbations onto clean images. The methods for generating AEs can be roughly classified into three categories based on the information of victim DNNs available to the attacker. The first category is the white-box attack \cite{dong2018boosting,croce2020minimally,zhang2020walking}, in which the attack method has access to the architecture and weights of the victim model. The second category is called the black-box attack, where the attacker can only obtain the model's output. The case where the model's output contains both categories and scores is known as a score-based black-box attack, while the situation where only categories are included is named a decision-based black-box attack \cite{brendel2018decision}. The third category refers to the gray-box attack \cite{guo2018countering}, where attackers can obtain models' knowledge that falls between white-box attacks and black-box attacks. Given that contemporary potent classification APIs \cite{GoogleCloud,Imagga} generally provide image prediction categories alongside scores, this work primarily focuses on practical score-based black-box attacks.

We observe that current state-of-the-art (SOTA) score-based black-box attack methods predominantly exhibit two aspects that can be improved. On the one hand, prior techniques that seek AEs by optimizing image features usually operate within the entire feature space, such as AdvFlow \cite{mohaghegh2020advflow}, TREMBA \cite{huangblack} and CGA \cite{feng2022boosting}, without embracing the disentanglement of adversarial capability and visual characteristics in the image feature. Optimizing the entire feature of images perhaps results in substantial pixel perturbations that will encounter truncation by the predefined disturbance constraint, thus diminishing the optimization efficacy of the features and eventually increasing query numbers. Another possible chain reaction is the decrease in the Attack Success Rate (ASR) caused by the limited query numbers. 

To improve the unsatisfactory ASR and query efficiency, our DifAttack disentangles an image's latent feature into an adversarial feature and a visual feature, where the former dominates the adversarial capability of an image, while the latter largely determines its visual appearance. Such a disentanglement is feasible because the adversarial capability is mainly entwined with the decision boundary of classifiers and is relatively independent of the intrinsic image signal, whereas a contrast holds true for the visual appearance. Afterward, in DifAttack, the adversarial feature is iteratively optimized according to the query feedback from the victim model, while keeping the visual feature unaltered, until a successful AE can be reconstructed from the fusion of these two features.

On the other hand, some score-based attack techniques including P-RGF \cite{cheng2019improving} and GFCS \cite{lord2022attacking}, exploit the gradients from local white-box surrogate models to update AEs during each query. These methods are effective when image categories of the training dataset for surrogate models and the victim model are the same, namely a \textbf{close-set} black-box attack scenario analogy to the close-set recognition setting \cite{vaze2021open}. When it comes to an \textbf{open-set} black-box attack, wherein the training dataset of the victim model is unknown, the effectiveness of these methods declines, as the possible incongruence between the output categories of surrogate models and the victim model hampers the calculation of gradients from surrogate models. Although there are other approaches, such as SimBA \cite{guo2019simple} and CGA \cite{feng2022boosting}, which are workable in both the open-set and close-set scenarios. These methods still face challenges in terms of query efficiency and ASR, especially in the challenging targeted attack. Our DifAttack circumvents the need to compute gradients of surrogate models when updating AEs to be queried. Instead, before embarking upon querying the victim model, we leverage AEs generated by performing white-box attack methods on surrogate models to learn how to extract a disentangled adversarial feature and visual feature from the image's latent feature, which enables DifAttack to more effectively conduct black-box attacks in the open-set scenarios. The efficiency of DifAttack in the open-set scenarios can be attributed to the fact that the disentangled feature space for optimizing AEs is also responsible for the image reconstruction task, which means that an image can still be reconstructed into its original form with minimal loss after being mapped into our disentangled feature space. In contrast, the feature space in other methods, such as TREMBA and CGA, is directly used to generate adversarial perturbations that are tightly bound to the data distribution of the training dataset. The latent feature for image reconstruction in DifAttack has better generalizability on unknown datasets than the feature space learned from adversarial perturbation distributions of a specific dataset. 

Specifically, in our DifAttack, we firstly train an autoencoder equipped with our proposed \textbf{D}ecouple-\textbf{F}usion (\textbf{DF}) module to disentangle adversarial features as well as visual features and to realize image reconstructions from these features. Then, in the attack stage, we iteratively update the adversarial feature using the natural evolution strategy \cite{wierstra2014natural}, eventually reconstructing an AE from the fusion of the perturbed adversarial feature and the clean image's visual feature. To the best of our knowledge, this may be the first work that employs the disentangled adversarial and visual features for black-box adversarial attacks. The contributions of our work can be summarized as follows:

 \begin{itemize}
 
 \item We design a new disentanglement method to distill an adversarial feature, which has a significant impact on
the image’s adversarial capability while minimizing its influence on visual perception, and a visual feature that has the opposite attributes, from an image's latent feature.

		\item Based on this disentanglement, we propose a new black-box attack method, DifAttack, which generates AEs by optimizing the adversarial feature according to the query feedback while maintaining the visual feature invariant. 
		
		\item Experimental results demonstrate that our approach outperforms SOTA score-based black-box attack methods in ASR and query numbers simultaneously in almost all cases, especially in targeted attacks and open-set scenarios. DifAttack also achieves perfect (100\%) ASR in most datasets and reduces query numbers by over 40\%. 
  
 \end{itemize}

 \begin{figure*}[h]
\centering
\includegraphics[width=0.65\textwidth]{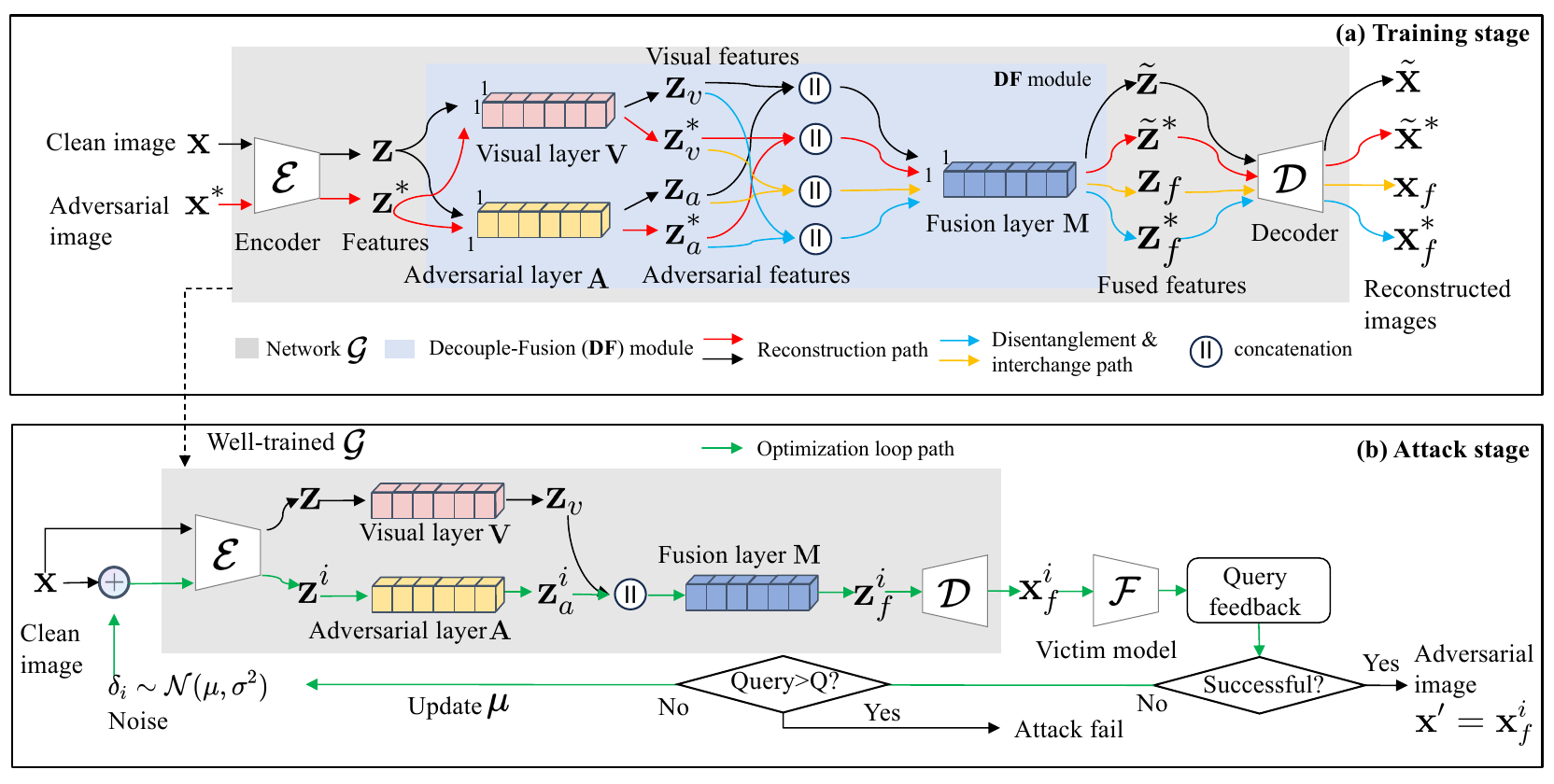} 
\caption{(a) The training procedure of the autoencoder $\mathcal{G}$ equipped with our proposed DF module for disentangling adversarial and visual features. (b) The proposed black-box adversarial attack method, i.e. DifAttack, incorporated with the pre-trained $\mathcal{G}$.}
\label{fig1}
\end{figure*}

\section{Related Works}

\subsection{Score-based Black-box Attack Methods}

Score-based black-box attack methods can be broadly categorized into three main classes: \textbf{transfer-based}, \textbf{query-based}, and \textbf{query-and-transfer-based} attacks \cite{feng2022boosting}.
1) \textbf{Transfer-based} approaches query the black-box model once using AEs generated on local surrogate models, as explored in \cite{qin2023training} and etc. Such approaches take advantage of the transferability of AEs across classifiers with distinct architectures, but their ASRs generally fall behind query-based attack methods. 
2) The \textbf{query-based} techniques tackle the black-box optimization through iteratively querying the victim model, including NES \cite{ilyas2018black}, Signhunter \cite{al2020sign}, $\mathcal{N}$Attack \cite{li2019nattack}, SimBA \cite{guo2019simple}, etc., which exhibit an augmented ASR compared to transfer-based approaches, albeit demanding a greater number of queries. 
3) The \textbf{query-and-transfer-based} methods amalgamate the merits of transfer-based and query-based techniques, resulting in an elevated ASR and diminished queries. Certain techniques, such as Subspace \cite{guo2019subspace}, P-RGF \cite{cheng2019improving}, GFCS \cite{lord2022attacking} and BASES \cite{cai2022blackbox}, update queried AEs by performing white-box attacks against surrogate models with the guidance of the query feedback. Other methods learn adversarial distributions by leveraging pairs of clean images and their AEs obtained by attacking surrogate models, including TREMBA \cite{huangblack} and CGA \cite{feng2022boosting}, or approximate the adversarial distribution with the clean data distribution, such as AdvFlow \cite{mohaghegh2020advflow}. 
Then, the learned feature space for mapping the adversarial distribution is adopted as the search space of AEs. Among them, CGA exhibits a remarkable improvement in query efficiency. However, its query number for targeted attacks on ImageNet still reaches over 3,000, leaving room for further improvement. 
Some other approaches incessantly refine an auxiliary model using the query feedback to directly generate AEs \cite{10017370}, approximate decision boundaries \cite{yin2023generalizable} or gradients \cite{duquery} of victim models, which are time-consuming and resource-intensive.

\subsection{Feature Disentanglement}

A lot of recent endeavors \cite{lin2019exploring,zou2020joint,wang2020cross} in computer vision have embraced the utilization of autoencoders \cite{hinton1993autoencoders,li2023comprehensive} and generative adversarial networks \cite{goodfellow2014generative,arora2023generative} to acquire disentangled representations of image features. In the domain of adversarial attacks and defenses, the disentanglement of features is commonly harnessed for detecting AEs and improving models' robustness, such as \cite{mustafa2020deeply,wang2021defending,yang2021adversarial}. To the best of our knowledge, only a limited number of \textit{white-box} attack methods have involved feature disentanglement. RNF \cite{kim2021distilling} is a white-box attack approach that optimizes image perturbations by maximizing the gradient of non-robust features in perturbed images, wherein non-robust features are distilled from the target classifier's intermediate features based on information bottleneck \cite{alemi2016deep}. Another white-box attack method \cite{lu2021discriminator} trains an autoencoder, called SSAE, to directly generate target models' AEs. Its training loss aims to minimize the norm difference of output logits between clean images and AEs to achieve better visual consistency meanwhile maximizing the angle discrepancy of output logits for mis-classification. We conjecture that the logits of the deep penultimate layer overfit to the specific model architecture, which makes it lack strong adversarial transferability in a black-box attack scenario. Compared to SSAE and RNF, our DifAttack decouples features related to the adversarial capability and visual perception from latent features used for image reconstructions, not from the intermediate features of specific classifiers, which is probably more suitable for black-box adversarial attacks.

\section{Proposed DifAttack}
\subsection{Method Overview}
As illustrated in Figure \ref{fig1}, our approach unfolds in two phases. In the first \textit{training stage}, we train an autoencoder $\mathcal{G}$ for feature disentanglement and image reconstruction, by using pairs of clean image $\mathbf{x}$ along with its AE $\mathbf{x}^*$ generated by performing untargeted white-box attacks on surrogate models. Taking the clean image $\mathbf{x}$ as an example, $\mathcal{G}$ can disentangle the latent feature $\mathbf{z}$ of $\mathbf{x}$ into the adversarial feature $\mathbf{z}_a$ and visual feature $\mathbf{z}_v$ via our proposed DF module, and then reconstructs another image $\tilde{\mathbf{x}}$ similar to $\mathbf{x}$ by decoding the fused $\mathbf{z}_a$ and $\mathbf{z}_v$. In the second \textit{attack stage}, we leverage the pre-trained $\mathcal{G}$ to generate an AE $\mathbf{x}^\prime$ of $\mathbf{x}$ to attack the black-box model $\mathcal{F}$. This $\mathbf{x}^\prime$ is obtained through an iterative optimization process, wherein we continuously sample a batch of perturbed images to extract adversarial features $\mathbf{z}^i_a$ by updating the sampling parameters according to the query feedback, while keeping the visual feature $\mathbf{z}_v$ consistent with that of the clean image, resulting in batches of reconstructed images $\mathbf{x}^i_f$ to query $\mathcal{F}$, until either the query number reaches its maximum threshold $Q$ or one of the perturbed images $\mathbf{x}^i_f$ successfully deceives $\mathcal{F}$, thereby yielding the AE $\mathbf{x}^\prime=\mathbf{x}_f^i$.

\subsection{Training Stage: Train the Autoencoder $\mathcal{G}$}
Suppose that there is a clean image $\mathbf{x}$ in the training dataset and we have pre-trained several white-box surrogate models on this dataset, forming a set $\mathbb{C}$. Note that the availability of these white-box surrogate models has also been assumed in the existing approaches \cite{lord2022attacking,feng2022boosting}. We adopt the white-box attack method PGD \cite{madry2018towards} to generate an AE $\mathbf{x^*}$ of $\mathbf{x}$ against a surrogate model $\mathcal{C}_j$ selected randomly from $\mathbb{C}$. PGD can be replaced by other white-box attack methods or a collection of them, in which case the performance is presented in the supplementary file. We aim to train an autoencoder $\mathcal{G}$ based on pairs of clean image $\mathbf{x}$ and its AE $\mathbf{x^*}$, for accomplishing two tasks: the feature disentanglement and image reconstruction. Next, we delve into the details of each task.
\subsubsection{Disentanglement.}For discovering an image's adversarial feature to which the image's adversarial capability is sensitive, meanwhile ensuring that alterations in this feature maintain minimal impact on the image's visual appearance, we disentangle adversarial and visual features from the latent feature used for image reconstruction. For realizing such a disentanglement, we exchange the adversarial and visual features between a pair of clean image and its AE, expecting images reconstructed from these exchanged features to exhibit different adversarial capabilities and visual appearance according to the original images their features stem from.

Specifically, for implementing the feature disentanglement, as illustrated in Figure \ref{fig1}(a), we design a DF module to not only extract but also fuse the adversarial and visual features. The DF module consists of the adversarial layer $\mathbf{A}$, visual layer $\mathbf{V}$ and fusion layer $\mathbf{M}$, which are all realized with a series of stacked $1\times1$ convolution layers. Taking the clean image $\mathbf{x}$ as an example, DF feeds its latent feature $\mathbf{z}=\mathcal{E}(\mathbf{x})$, where $\mathcal{E}$ denotes a Convolutional Neural Network (CNN)-based encoder, into $\mathbf{A}$ and $\mathbf{V}$, then outputs a new feature $\tilde{\mathbf{z}}$ from $\mathbf{M}$. The layers $\mathbf{A}$ and $\mathbf{V}$ employ distinct weights to activate $\mathbf{z}$ into the adversarial feature $\mathbf{z}_a$ and the visual feature $\mathbf{z}_v$, respectively. The $\tilde{\mathbf{z}}$ is capable of reconstructing $\Tilde{\mathbf{x}}\approx\mathbf{x}$ through the CNN-based decoder $\mathcal{D}$. We formulate the functionality realized by the DF module as:
\begin{small}
\begin{equation}\label{eq:defu1}
\begin{aligned}
\mathrm{DF}(\mathbf{z},\mathbf{z})=\mathbf{M}(\mathbf{A}(\mathbf{z})||\mathbf{V}(\mathbf{z})).
\end{aligned}
\end{equation}
\end{small}Here, $||$ denotes the channel-wise concatenation. Notably, DF does not satisfy the commutative property, i.e. $\mathrm{DF}(\mathbf{z}^*,\mathbf{z})\neq\mathrm{DF}(\mathbf{z},\mathbf{z}^*)$, where $\mathbf{z}^*=\mathcal{E}(\mathbf{x^*})$, unless $\mathbf{z}=\mathbf{z}^*$. 

Then we utilize the DF module to conduct feature interchange for accomplishing disentanglement. Specifically, as depicted by the yellow path in Figure \ref{fig1}(a), the adversarial feature of the clean sample, i.e. $\mathbf{z}_a$, is firstly concatenated with the visual feature of the adversarial image, i.e. $\mathbf{z}_v^*$. Afterward, they are inputted into the fusion layer to obtain the fused feature $\mathbf{z}_f$, thereafter decoded into a new image $\mathbf{x}_f$. Symmetrically, as shown by the blue path, we can reconstruct another image $\mathbf{x}_f^*$ from $\mathbf{z}_a^*=\mathbf{A}(\mathbf{z}^*)$ and $\mathbf{z}_v=\mathbf{V}(\mathbf{z})$. The generation of $\mathbf{x}_f$ and $\mathbf{x}_f^*$ can be formulated as:

\begin{equation}\label{eq:reconx}
\begin{small}
\begin{gathered}
\mathbf{x}_f=\mathcal{D}(\mathrm{DF}(\mathbf{z},\mathbf{z}^*) ),\
\mathbf{x}_f^*=\mathcal{D}(\mathrm{DF}(\mathbf{z}^*,\mathbf{z}) ).
\end{gathered}
\end{small}
\end{equation} 
Subsequently, since we desire the adversarial feature to control the adversarial capability of the image, while the visual feature to govern its visual perception, $\mathbf{x}_f$ should be closer to the visual perception of the AE $\mathbf{x^*}$ and be non-adversarial. Conversely, $\mathbf{x}_f^*$ is expected to be closer to the visual perception of the clean image $\mathbf{x}$ and be adversarial against all surrogate models so as to improve the generalization ability of the adversarial feature on different classifiers. Hence, we aim to minimize the disentanglement loss ${L}_{dis}$ defined as:
\begin{equation}\label{eq:disenloss}
\begin{small}
\begin{gathered}
{L}_{dis} = ||\mathbf{x}^*-\mathbf{x}_f||_2+\frac{1}{N}\sum\nolimits_{j=1}^{N}{L}_{adv}(\mathbf{x}_f,y,1,\mathcal{C}_j,k)\\
+||\mathbf{x}-\mathbf{x}_f^*||_2+\frac{1}{N}\sum\nolimits_{j=1}^{N}{L}_{adv}(\mathbf{x}_f^*,y,0,\mathcal{C}_j,k),
\end{gathered}
\end{small}
\end{equation}
where $N$ signifies the cardinality of the surrogate model set $\mathbb{C}$. The $l_2$ norm measures the Euclidean distance of two images, and ${L}_{adv}$ is employed to evaluate the adversarial ability, whose definition is \cite{carlini2017towards}:
\begin{equation}
\begin{small}
\begin{gathered}\label{eq:advloss}
{L}_{adv}(\mathbf{x},y,v,\mathcal{C}_j,k) \\ 
=\max\left\{\mathbb{I}(v)\cdot\big(\mathcal{C}_j(\mathbf{x},y)-\max\limits_{d\neq y}\mathcal{C}_j(\mathbf{x},d)\big),-k\right\}.
\end{gathered}
\end{small}
\end{equation}
 The value of $v$ can be set as 0 or 1. $\mathbb{I}(0)=1$ means an untargeted attack with a ground-truth label of $y$, while $\mathbb{I}(1)=-1$ indicates a targeted attack with a target category of $y$. $\mathcal{C}_j(\mathbf{x},d)$ is the output score of $\mathcal{C}_j$ at the $d$-th class. The adjustable constant $k>0$ controls the adversarial extent.

\subsubsection{Reconstruction.} 
For achieving a cycle consistency \cite{zhu2017unpaired} with regard to feature disentanglement, as depicted in the reconstruction paths of Figure \ref{fig1}(a), the autoencoder is expected to reconstruct an input image well even when its latent feature is processed by the DF module. Given that the image decoupled by $\mathcal{G}$ could be either clean or adversarial, we define the reconstruction loss ${L}_{rec}$ of $\mathcal{G}$ by:
\begin{equation}\label{eq:reconloss}
\begin{small}
\begin{aligned}
{L}_{rec} = ||\mathbf{x}-\mathcal{D}\big(\mathrm{DF}(\mathbf{z},\mathbf{z})\big)||_2
+||\mathbf{x^*}-\mathcal{D}\big(\mathrm{DF}(\mathbf{z}^*,\mathbf{z}^*)\big)||_2.
\end{aligned}
\end{small}
\end{equation}
Eventually, the loss function to be minimized during training the network $\mathcal{G}$ can be computed as:
\begin{equation}\begin{gathered}\label{eq:totalloss}
{L}_{all} =\lambda\cdot{L}_{rec} +{L}_{dis},
\end{gathered}
\end{equation}
where the hyper-parameter $\lambda$ balances two loss terms.

\begin{algorithm}[tb]
\caption{The proposed DifAttack method.}
\label{alg:algorithm}
\textbf{Input}: target classifier $\mathcal{F}$; clean image $\mathbf{x}$; max query budget $Q>0$; distortion budget $\epsilon$; ground-truth label $y$; learning rate $\eta$; variance $\sigma$; $v$, $k$ in Eq.(\ref{eq:algo3}); sample scale $\tau$.\\ 
\textbf{Output}: Adversarial image $\mathbf{x}^\prime$.
\begin{algorithmic}[1] 
\STATE Let $q\gets0,\mathbf{x}^\prime\gets\mathbf{x},\mathbf{z}\gets\mathcal{E}(\mathbf{x}),\mu\sim\mathcal{N}(0,I)$
\WHILE{$q+\tau\leq Q$}
\STATE  $\gamma_1,...,\gamma_i,...,\gamma_\tau \sim\mathcal{N}(0,I)$
\STATE $\delta_i\gets \mu+\sigma\gamma_i\,,\forall i\in\{1,...,\tau\}$
\STATE $\mathbf{x}_f^i\gets\Pi_{\epsilon,\mathbf{x}}(\mathcal{D}(\mathrm{DF}(\mathcal{E}(\mathbf{x}+\delta_i ),\mathbf{z})))\,,\forall i$
\STATE  $l_i \gets {L}_{adv}(\mathbf{x}_f^i,y,v,\mathcal{F},k)\,,\forall i$
\STATE $q\gets q+\tau$
\IF {$\exists\,l_i=-k$}
\STATE  $\mathbf{x^\prime}\gets\mathbf{x}_f^i$
\STATE \textbf{break}
\ELSE
\STATE $\hat{l}_i\gets(l_i-$mean$(\{l_i\}))/$std$(\{l_i\})\,,\forall i$
\STATE $\mu\gets\mu-\frac{\eta}{\tau\sigma}\sum_{i=1}^{\tau}\hat{l_i}\gamma_i$
\ENDIF
\ENDWHILE
\STATE \textbf{return} $\mathbf{x^\prime}$
\end{algorithmic}
\end{algorithm}

\begin{table*}[htbp]
  \centering
  
  \scalebox{0.65}{
    \begin{tabular}{c|cc|cc|cc|cc|cc|cc|cc|cc}
    \hline
       Datasets   & \multicolumn{8}{c|}{ImageNet} & \multicolumn{8}{c}{Cifar-10}                                  \Tstrut\\
    \hline
    Victim Models & \multicolumn{2}{c|}{ResNet} & \multicolumn{2}{c|}{VGG} & \multicolumn{2}{c|}{GoogleNet} & \multicolumn{2}{c|}{SqueezeNet} & \multicolumn{2}{c|}{PyramidNet} & \multicolumn{2}{c|}{DenseNet} & \multicolumn{2}{c|}{VGG} & \multicolumn{2}{c}{ResNet} \Tstrut \\
    \hline
    Methods & ASR   & Avg.Q   & ASR   & Avg.Q   & ASR   & Avg.Q   & ASR   & Avg.Q   & ASR   &  Avg.Q  & ASR   & Avg.Q   & ASR   & Avg.Q   & ASR   & Avg.Q \Tstrut\\
    \hline
    Signhunter &        49.8  &        7,935  &               60.8  &     7,606  &            28.1  &        9,021  &          45.7  &       8,315 &\textbf{     100.0 } &          1,193  & \textbf{      100.0 } &        1,364  &            99.6  &        1,792  &          99.9  &          1,137      \Tstrut\\
    NES    &          54.3  &        7,988  &               47.7  &     8,328  &            37.7  &        8,748  &          45.7  &       8,235  &          96.3  &          1,742  &          98.0  &        1,757  &            85.8  &        3,300  &          96.1  &          1,745 \Tstrut\\
    $\mathcal{N}$Attack  &          94.5  &        5,324  &               93.5  &     5,924  &            76.9  &        6,883  &          90.0  &       5,644 & \textbf{     100.0 } &          1,072  & \textbf{      100.0 } &        1,185  & \textbf{        100.0 } &        1,983  & \textbf{     100.0 } &          1,168   \Tstrut\\
    SimBA&          97.0  &        3,873  &               97.5  &     3,505  &            77.5  &        5,874  &          94.0  &       4,230  & \textbf{     100.0 } &             \underline{812}  & \textbf{      100.0 } &           888  &            99.4  &        \underline{1,182}  & \textbf{     100.0 } &             \underline{824}   \Tstrut\\
    BASES  &          35.3  &        7,170  &               26.7  &     8,229  &            12.5  &        8,901  &          15.4  &       9,228 &  38.5     &    6,412   &  29.5     &  7,433     &  37.5     &  6,682     &  44.5     &   5,990     \Tstrut\\
    P-RGF &          51.3  &        6,393  &               58.3  &     6,093  &            34.2  &        7,790  &          65.3  &       5,433  &          94.2  &          1,297  &          97.7  &           742  &            89.6  &        1,372  &          91.6  &          1,546  \Tstrut\\
    Subspace  &          66.3  &        4,750  &               68.3  &     4,964  &            50.3  &        6,642  &          33.2  &       8,063 &          88.8  &          3,035  &          87.0  &        2,490  &            85.1  &        2,816  &          80.1  &          3,603  \Tstrut\\
    GFCS  &          95.0  &        \underline{3,274}  &               93.0  &     \underline{3,358}  &            60.0  &        6,359  &          72.5  &       6,127  &          96.9  &          1,045  &          99.3  &           \underline{635}  &            93.9  &        1,505  &          97.1  &             979  \Tstrut\\
    CGA  &          91.6  &        4,222  &               91.1  &     4,801  &            90.9  & \underline{4,645} &          93.2  &       \underline{3,972} & 98.5  & 1,759 & 99.4  & 1,953 & 91.0    & 2,673 & 99.0    & 1,309 \Tstrut\\
    \hline
    DifAttack(\textbf{Ours}) & \textbf{      100.0 } & \textbf{     1,901 } & \textbf{           100.0 } & \textbf{   2,246 } & \textbf{          97.0 } & \textbf{      3,201 } & \textbf{        98.0 } & \textbf{     2,793 } & \textbf{     100.0 } & \textbf{           414 } & \textbf{      100.0 } & \textbf{        498 } & \textbf{        100.0 } & \textbf{         736 } & \textbf{     100.0 } & \textbf{           455 }  \Tstrut\\
     
    \hline
    \end{tabular}%
    }
    \caption{The attack success rate (ASR \%), average number of queries (Avg.Q) of test images in targeted attacks with the target class 864 for ImageNet and the target class 0 for Cifar-10. The best and second best Avg.Q are \textbf{bold} and \underline{underlined}, respectively.}
  \label{tab:targeted}%
\end{table*}%

\subsection{Attack Stage: Generate the AE $\mathbf{x}^\prime$} 
 Upon obtaining a well-trained network $\mathcal{G}$, in the subsequent attack stage, we can utilize it to generate AEs in a black-box setting. As shown in Figure \ref{fig1}(b), we maintain the visual feature for decoding the perturbed image $\mathbf{x}_f^i$ to be identical to that of the clean image, i.e. $\mathbf{z}_v$. This ensures that $\mathbf{x}_f^i$ remains in close proximity to the clean image. Meanwhile, our objective is to identify an adversarial feature that, when combined with $\mathbf{z}_v$, has the ability to deceive the victim model $\mathcal{F}$. Therefore, we aim to discover a perturbation $\delta$ whose corresponding perturbed image $\mathbf{x}+\delta$ can be used to extract such an adversarial feature. Then the final adversarial image $\mathbf{x}^\prime$ can be reconstructed as follows:

 \begin{equation}\begin{small}\begin{gathered}\label{eq:adv1}
\mathbf{x}^\prime=\Pi_{\epsilon,\mathbf{x}}\bigg(\mathcal{D}\Big(\mathrm{DF}\big(
\mathcal{E}(\mathbf{x}+\delta),\mathcal{E}(\mathbf{x})\big)\Big)\bigg),
\end{gathered} 
\end{small}\end{equation}
where the operation $\Pi_{\epsilon,\mathbf{x}}$ represents the projection of the input image into the $l_p$ ball bounded by $\epsilon$ and centered at $\mathbf{x}$. This forces the adversarial perturbation to remain within the specified constraint. For simplicity, we recast the right-hand side of Eq.(\ref{eq:adv1}) as $\mathbf{T}(\mathbf{x}+\delta)$. To find the AE $\mathbf{x}^\prime$, by combining Eq.(\ref{eq:adv1}) and using the transformation of variable approach, we formulate our attack objective in a unified manner as:
\begin{equation}\begin{small}\begin{gathered}\label{eq:algo3}
\mathrm\min_{\mu}\mathbb{E}_{\mathcal{N}(\delta|\mu,\sigma^2)}{L}_{adv}(\mathbf{T}(\mathbf{x}+\delta),y,v,\mathcal{F},k),\\
s.t. \ ||\mathbf{T}(\mathbf{x}+\delta)-\mathbf{x}||\leq \epsilon,
\end{gathered}\end{small}
\end{equation}
 where $\mathbf{x}^\prime=\mathbf{T}(\mathbf{x}+\delta)$ satisfies $\mathrm{\arg\max_{d}}\mathcal{F}(\mathbf{x}^\prime,d)\neq y$ for untargeted attacks or $\mathrm{\arg\max_{d}}\mathcal{F}(\mathbf{x}^\prime,d)= t$ for targeted attacks. Here, $t$ denotes the target class.
We sample $\delta$ from a Gaussian distribution $\mathcal{N}(\mu,\sigma^2)$ where the mean $\mu$ is to be optimized and the variance $\sigma$ is fixed to be the optimal value through performing a grid search. For optimizing $\mu$, we adopt the natural evolution strategies method to estimate the gradient of the expectation of ${L}_{adv}$ in Eq.(\ref{eq:algo3}) w.r.t $\mu$. When using stochastic gradient descent with a learning rate $\eta$ and a batch size $\tau$, the update rule for $\mu$ is as follows:
\begin{equation}\begin{small}\begin{gathered}\label{eq:algo4}
\mu \gets \mu- \frac{\eta}{\tau\sigma}\sum\nolimits_{i=1}^{\tau}\gamma_i{L}_{adv}(\mathbf{T}(\mathbf{x}+\mu+\sigma\gamma_i),y,v,\mathcal{F},k).
\end{gathered}\end{small}
\end{equation}
Here, $\gamma_i\sim\mathcal{N}(0,I)$, and by setting $\delta_i=\mu+\sigma\gamma_i$, we obtain $\delta_i\sim\mathcal{N}(\mu,\sigma^2)$. 

Finally, our proposed DifAttack is summarized in Algorithm 1. Note that, in line 12, we normalize the $L_{adv}$ of each perturbed image to stabilize convergence by taking inspiration from the $\mathcal{N}$Attack \cite{li2019nattack} method.

\section{Experiments}

In this section, we compare our DifAttack with SOTA score-based black-box attack methods in both the close-set and open-set scenarios. Ablation experiments on the DF module and $\tau$ in Eq.(\ref{eq:algo4}) are also conducted. Further experimental results are presented in the supplementary file.

\subsection{Experiment Setup}

\subsubsection{Datasets.} We mainly conduct experiments on the large-scale ImageNet-1k, small-scale Cifar-10 and Cifar-100 datasets. The images in ImageNet are cropped and resized to $224\times224$, while images in Cifar-10 and Cifar-100 datasets remain a size of $32\times32$. When conducting attacks on Cifar-100 in open-set scenarios, we exclude 10 classes that have potential semantic overlap with Cifar-10. The specific classes are presented in the supplementary file.

\begin{table}[t]
  \centering
 
  \scalebox{0.65}{
    \begin{tabular}{c|cc|cc|cc|cc}
    \hline
   
    Victim Models & \multicolumn{2}{c|}{ResNet} & \multicolumn{2}{c|}{VGG} & \multicolumn{2}{c|}{GoogleNet} & \multicolumn{2}{c}{SqueezeNet} \Tstrut\\
    \hline
    Methods & ASR   & Avg.Q   & ASR   & Avg.Q   & ASR   & Avg.Q   & ASR   & Avg.Q    \Tstrut\\
    \hline
    Signhunter& 99.5  & 931   & \textbf{100.0} & 702   & 95.5  & 1,504 & \textbf{100.0} & 363    \Tstrut\\
    NES   & 97.5  & 1,264 & 98.0  & 1,068 & \textbf{100.0} & 1,234 & 91.0  & 1,711   \Tstrut\\
    $\mathcal{N}$Attck  & \textbf{100.0} & 780   & \textbf{100.0} & 752   & \textbf{100.0} & 878   & \textbf{100.0} & 563 \Tstrut\\
    SimBA  & 99.5  & 614   & 99.0  & 533   & \textbf{100.0} & 865   & 99.5  & 428  \Tstrut\\
    BASES  & 86.4  & 1,549 & 95.5  & 564   & 81.7  & 2,080 & 85.1  & 1,717  \Tstrut\\
    P-RGF & 96.0  & 711   & 98.0  & 529   & 93.0  & 1,162 & 95.0  & 796 \Tstrut\\
    Subspace  & 94.5  & 917 & 95.5  & 867   & 95.5  & 994   & 97.0  & 566  \Tstrut\\
    GFCS  & \textbf{100.0} & 739   & \textbf{100.0} & 515   & 96.5  & 1,348 & 98.5  & 658 \Tstrut\\
    CGA & 97.3  & \underline{475}   & 99.4  & \underline{137}   & \textbf{100.0} & \textbf{139}   & 99.3  & \underline{202} \Tstrut\\
    \hline
    DifAttack(\textbf{Ours})  & \textbf{100.0} & \textbf{156} & \textbf{100.0} & \textbf{123}   & \textbf{100.0} & \underline{165}   & \textbf{100.0} & \textbf{90}  \Tstrut\\

    \hline
    \end{tabular}%
    }
     \caption{The ASR (\%), Avg.Q of test images from ImageNet in an untargeted attack setting.}
  \label{tab:untargetedImageNet}%
\end{table}%

\subsubsection{Classifiers.} We collect four pre-trained classifiers with different architectures for each of the above three datasets. In the close-set scenario, when we attack one of the four models, the other three are adopted as surrogate models. When attacking the model of Cifar-100 in the open-set scenario, the surrogate models consist of three classifiers trained on Cifar-10 with different architectures from the victim model. For ImageNet, the four models are ResNet-18, VGG-16, GoogleNet, and SqueezeNet provided by torchvision\footnote[2]{https://pytorch.org/vision/stable/index.html}. For Cifar-10, ResNet-Preact-100, VGG-15, DenseNet-BC-110 and PyramidNet-110 \cite{feng2022boosting} are selected. For Cifar-100, its four classifiers are chosen from ResNet-56, VGG-16, DenseNet-121 and GoogleNet \cite{PytorchCifar100}.

\subsubsection{Comparative Methods.} We compare our DifAttack with SOTA approaches, including Signhunter \cite{al2020sign}, NES \cite{ilyas2018black}, $\mathcal{N}$Attack \cite{li2019nattack}, SimBA-DCT (shorted as SimBA) \cite{guo2019simple}, BASES \cite{cai2022blackbox}, P-RGF \cite{cheng2019improving}, Subspace \cite{guo2019subspace}, GFCS \cite{lord2022attacking} and CGA \cite{feng2022boosting}. We evaluate them according to their official source codes. Due to the absence of pre-trained models and code, the experimental results of the CGA method on ImageNet and the real-world Imagga API \cite{Imagga} are cited from its published paper, while its results on Cifar-10 and Cifar-100 are derived from the pre-trained models provided by the authors.

\begin{table*}[htbp]
  \centering

  \scalebox{0.7}{
    \begin{threeparttable}
    \begin{tabular}{c|cc|cc|cc|cc|cc|cc|cc|cc}
    \hline
     Settings     & \multicolumn{8}{c|}{Untargeted}                               & \multicolumn{8}{c}{Targeted}  \Tstrut\\
    \hline
    Victim Models & \multicolumn{2}{c|}{GoogleNet } & \multicolumn{2}{c|}{DenseNet} & \multicolumn{2}{c|}{ResNet} & \multicolumn{2}{c|}{VGG} & \multicolumn{2}{c|}{GoogleNet} & \multicolumn{2}{c|}{DenseNet} & \multicolumn{2}{c|}{ResNet} & \multicolumn{2}{c}{VGG}  \Tstrut\\
    \hline
    Methods & ASR   & Avg.Q & ASR   & Avg.Q & ASR   & Avg.Q & ASR   & Avg.Q & ASR   & Avg.Q & ASR   & Avg.Q & ASR   & Avg.Q & ASR   & Avg.Q \Tstrut \\
    \hline
    Signhunter & 99.0  &           519  & 99.5  &           386  & \textbf{100.0} &          121  & 97.5  &           704  & 89.5  &          \underline{3,586}  & 92.0  &          \underline{3,336}  & 96.5  &          \underline{2,076}  & 81.5  &          \underline{4,013}   \Tstrut\\
    NES   & 92.5  &        1,442  & 94.0  &        1,454  & 96.0  &          914  & 88.0  &        1,961  & 48.5  &          6,724  & 53.0  &          6,488  & 76.5  &          4,396  & 38.5  &          7,642   \Tstrut\\
    $\mathcal{N}$Attack & \textbf{100.0} &           553  & \textbf{100.0} &           528  & \textbf{100.0} &          311  & 97.0  &           952  &
    73.0  &          6,101  & 82.0  &          5,646  & 95.0  &          2,853  & 76.5  &          5,235     \Tstrut\\
    Simba & \textbf{100.0} &           \underline{327}  & 99.5  &           389  & 98.0  &          360  & \textbf{98.5} &           \underline{605}  & 79.5  &          4,203  & 78.5  &          4,189  & 96.0  &          2,401  & 77.5  &          4,156  \Tstrut \\
    CGA   & 96.5  &           571  & 99.5  &          \underline{279}  & \textbf{100.0} &            \underline{59}  & 96.5  &           691  & 70.1  &          5,098  & 75.6  &          4,781  & 90.0  &          3,189  & 63.2 &          5,586  \Tstrut \\
    \hline
    DifAttack(\textbf{Ours})  & \textbf{100.0} & \textbf{         146 } & \textbf{100.0} & \textbf{         122 } & \textbf{100.0} & \textbf{          39 } & \textbf{98.5} & \textbf{         386 } & \textbf{91.0} & \textbf{        2,506 } & \textbf{94.0} & \textbf{        2,740 } & \textbf{97.0} & \textbf{        1,382 } & \textbf{86.0} & \textbf{        2,980 } \Tstrut\\
    \hline
    \end{tabular}%
             \begin{tablenotes}   
        \footnotesize           
        \item[1] Methods including BASES, P-RGF, Subspace, and GFCS, which rely on the output scores of surrogate models to update perturbed images, are unable to perform attacks in this open-set scenario. The reason is that the categories outputted by these models do not encompass the classes of the test images from Cifar-100.          
      \end{tablenotes}            
    \end{threeparttable}       
    }
      \caption{Untargeted and targeted black-box attacks in the open-set scenario. The test images are sourced from CIFAR-100, and the victim model is trained on CIFAR-100. Meanwhile, the autoencoder and surrogate models are trained on CIFAR-10.}
  \label{tab:cifar100}%
\end{table*}%
\subsubsection{Parameters.} Following many previous works, we also set the maximum perturbation on ImageNet to 12/255, and on CIFAR-10 as well as CIFAR-100 to 8/255. The maximum query number for these three datasets is set to 10,000. For the real-world Imagga API, the maximum query number is limited to 500 due to their query limit. In our DifAttack, we set $\lambda=1$, $\sigma=0.1$, $\eta=0.01$, $k=5$ or $0$ for Eq.(\ref{eq:disenloss}) and (\ref{eq:algo4}).

\subsection{The Close-set Scenario}
\subsubsection{Black-box Attack on ImageNet and Cifar-10.}

Table \ref{tab:targeted} presents the ASR and average query numbers (Avg.Q) in challenging targeted attacks on ImageNet and Cifar-10. We randomly select a target class and 1,000 images, excluding those belonging to the selected class, as the test set. It is observed that our method achieves a 100\% ASR in most cases, which is much better than other approaches when images come from ImageNet. Additionally, our DifAttack significantly reduces Avg.Q when obtaining the same or even higher ASR compared to other competitors. For instance, when attacking ResNet on ImageNet, the Avg.Q of DifAttack is reduced by 41.9\% compared to the second-lowest query numbers obtained by GFCS (from 3,274 to 1,901), meanwhile achieving an additional 5\% increase in ASR. In apart from this, for Cifar-10, DifAttack also reduces the Avg.Q by 43.3\% compared to the second-lowest queries obtained by SimBA (from 812 to 414) when against PyramidNet. 
The experimental results of untargeted attacks on ImageNet are presented in Table \ref{tab:untargetedImageNet}, where DifAttack maintains a 100\% ASR in all cases and achieves lower query numbers in most cases. Given the inherent simplicity of the untargeted attack task, the Avg.Q of all competitors remains relatively low; but DifAttack is also almost the best. As presented earlier, in more challenging targeted attacks, DifAttack significantly outperforms other methods. 
Further experimental results for untargeted attacks on Cifar-10 and targeted attacks with other target classes can be found in supplementary file.

\begin{table}[t]
  \centering

  \scalebox{0.65}{
   \begin{threeparttable}
    \begin{tabular}{c|c|cccccc}
    \hline
    Settings & Metrics & DifAttack  & CGA & GFCS  & SimBA & $\mathcal{N}$Attack &Signhunter \Tstrut\\
    \hline
     \multirow{2}{*}{Untargeted} & ASR   & \textbf{90} & 85 & 65    & 45    & 65 & 70\Tstrut\\
       & Avg.Q & \textbf{90.8} & \underline{139.2} & 221.2 & 287.6 & 207.2 &187.2\Tstrut\\
    \hline
    \multirow{2}{*}{Targeted} & ASR   & \textbf{70}    & -  & -  &    35   & 30 &60\Tstrut\\
         & Avg.Q & \textbf{183.5} &   -    &    -   &      338.3 & 417.8&\underline{189.5} \Tstrut\\
    \hline
   
    \end{tabular}%
     \begin{tablenotes}   
        \footnotesize           
        \item[1] In untargeted attacks, we aim to remove the top three predicted classes of clean images. In targeted attacks, we set the target class to be the 2nd-highest scoring class among the API's predicted classes. 
      \end{tablenotes}  
    \end{threeparttable}   
    }
      \caption{The ASR and Avg.Q when attacking Imagga API.}
  \label{tab:imagga}%
\end{table}%

\subsection{The Open-set Scenario}
\subsubsection{Black-box Attack on Cifar-100.}

We use the autoencoder and surrogate models pre-trained on Cifar-10 to generate AEs for images from Cifar-100, then evaluate the performance of these AEs when attacking victim models pre-trained on Cifar-100. The experimental results in Table \ref{tab:cifar100} demonstrate that in both untargeted and targeted attacks, DifAttack can always achieve fewer query numbers and the same or higher ASR simultaneously, compared to other methods. For example, when against GoogleNet in an untargeted setting, DifAttack reduces the minimum Avg.Q for other methods from 327 to 146; in targeted attacks against VGG and GoogleNet, our approach decreases the lowest Avg.Q obtained by other methods by over 1,000 while maintaining a higher ASR.  
These experimental results in such an open-set scenario illustrate the favorable generalizability of our disentangled adversarial feature on unseen datasets.

 \begin{figure}[t]
\centering
\includegraphics[width=0.4\textwidth]{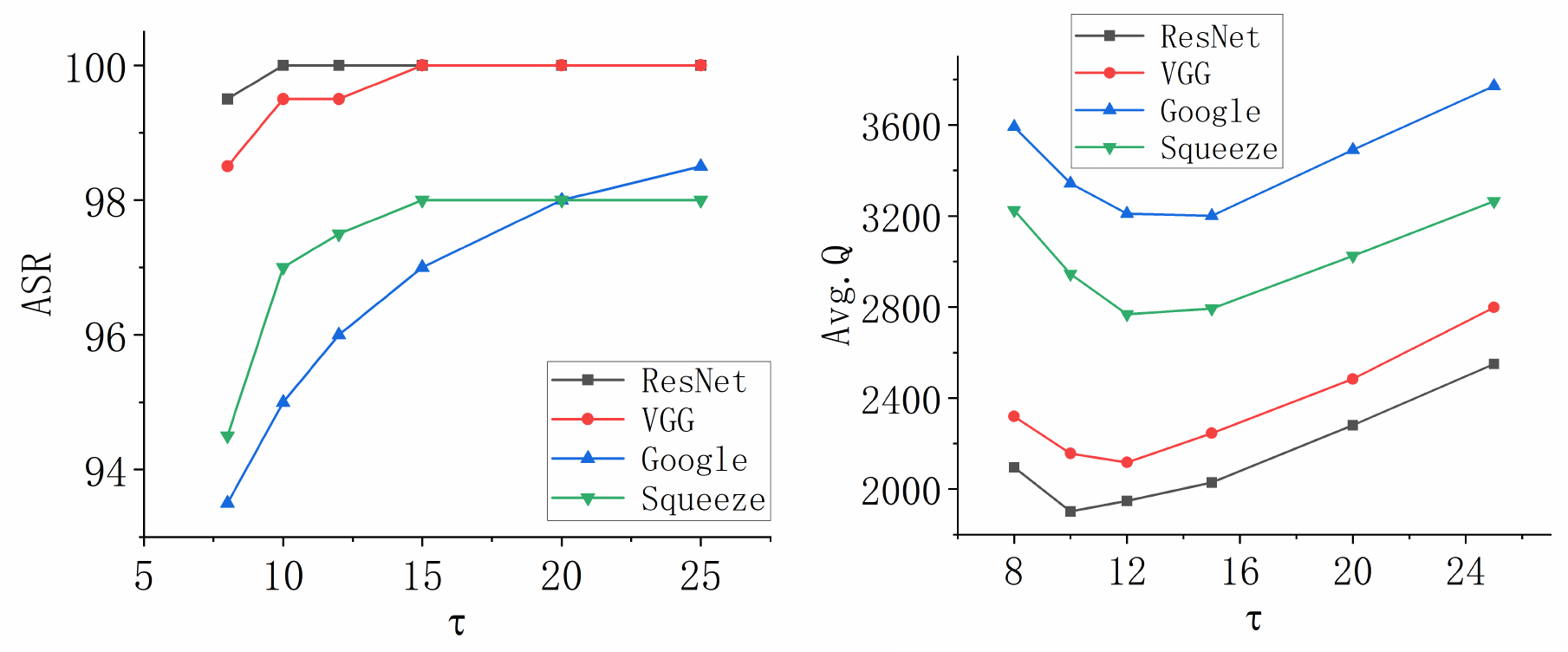} %
\caption{The influence of the parameter $\tau$ on the ASR and Avg.Q when performing targeted attacks on ImageNet.} 
\label{fig:featureSensitivityTargeted}
\end{figure}

 \begin{figure}[t]
\centering
\includegraphics[width=0.4\textwidth]{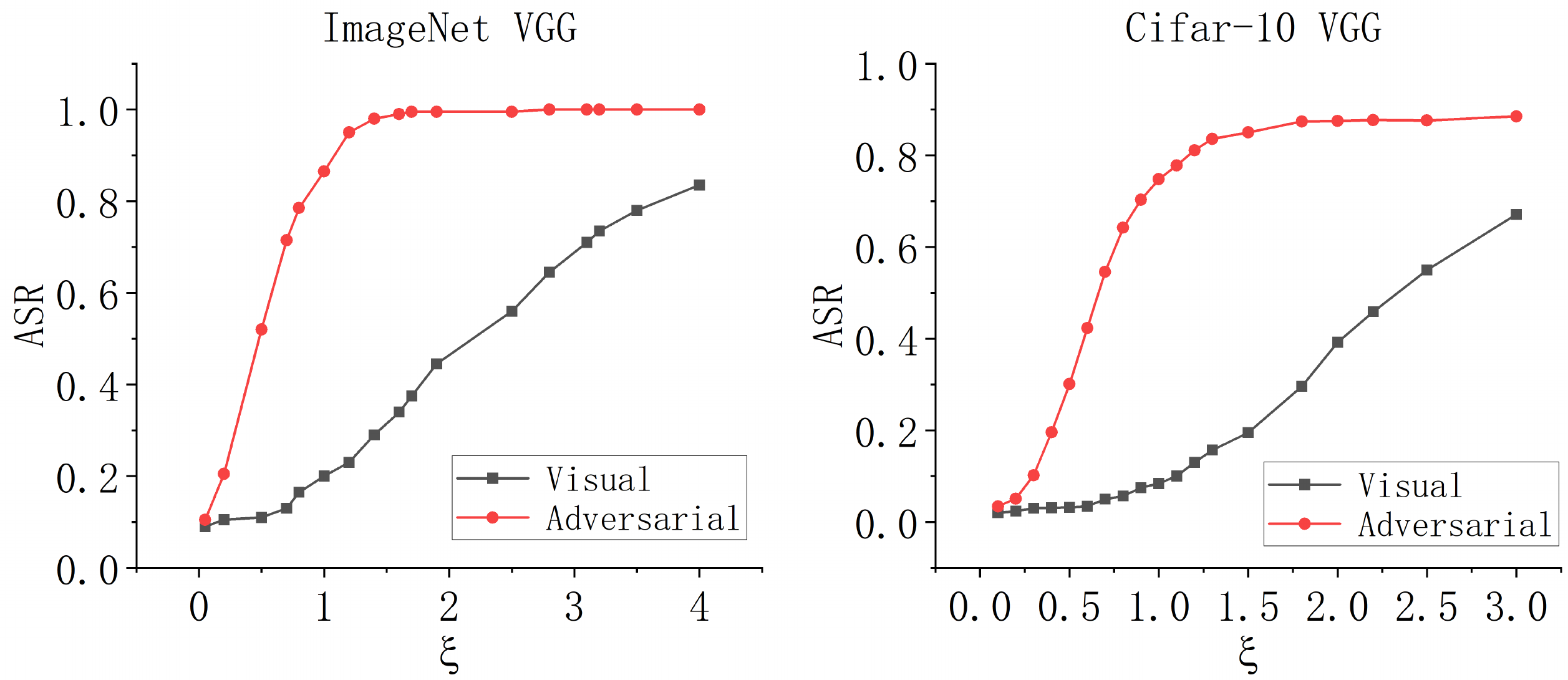} %
\caption{The sensitivity comparison of the ASR between the adversarial and visual features.} 
\label{fig:featureSensitivity}
\end{figure}

\subsubsection{Black-box Attack against Real-world API.}
To evaluate the attack performance in a more practical open-set scenario, we assess comparative methods by attacking the real-world Imagga API \cite{Imagga}. This API utilizes a recognition model trained on a diverse dataset of daily images spanning over 3,000 unknown categories. The Imagga provides multiple predicted classes and corresponding scores for each query image. Due to the API's call limits, we choose the top-performing five methods in the above experiments as the comparative methods, which are CGA, GFCS, SimBA, $\mathcal{N}$Attack and Signhunter. We use the autoencoder and surrogate models pre-trained on ImageNet to extract disentangled features of test images. The experimental results in Table \ref{tab:imagga} indicate that even when the actual training dataset is completely unknown, our DifAttack exhibits superior query efficiency and ASR, e.g., an Avg.Q of 90.8 for the untargeted attack, and 90\%/70\% ASR in untargeted/targeted attacks. 
We also observe that Signhunter, which has the second best targeted query efficiency in Table \ref{tab:cifar100}, achieves an Avg.Q comparable to ours in the targeted attack against Imagga, yet suffering an ASR that is 10\% lower than ours. Moreover, in the untargeted attack against Imagga and closed-set scenarios presented in Table \ref{tab:targeted}-\ref{tab:untargetedImageNet}, Signhunter largely lags behind our method in both ASR and Avg.Q. Notably, GFCS successfully generates AEs in untargeted attacks but not in targeted attacks, since GFCS is unable to craft AEs when the target class does not occur in the output of surrogate models. Apart from this, the missing result of CGA in the targeted attack is due to the absence of code and this result in their work.

\subsection{Black-box Attack against Defensive Methods}\label{app:defensive}

It is also important to evaluate the performance of attack methods against defensive strategies. We compare the ASR of comparative methods in untargeted attacks when encountering several SOTA defensive techniques, including Adversarially Training (AT) and pre-processing based methods, such as Bit-depth Reduction (Bit-Red) \cite{featureSqueeze}, Neural Representation Purifier (NRP) \cite{naseer2020self} and Diffusion Purifier (DiffPure) \cite{nie2022diffusion}. For attacking AT models pre-trained on ImageNet, we adopt adversarially trained ResNet-50 as the surrogate model and the more powerful classifier Wide-ResNet-50-2 as the victim model \cite{salman2020adversarially}. For other defensive techniques, we use them to process AEs from ImageNet generated for attacking standardly trained ResNet-18, and then evaluate the ASR of these AEs after being processed. The experimental results shown in Table \ref{tab:ATdefense} showcase that the average ASR of DifAttack surpasses that of other methods, standing as the sole approach with an average ASR exceeding 40\%. Additionally, across each defensive strategy, DifAttack achieves 2 second and 1 third places. The good performance indicates that DifAttack can effectively disentangle the adversarial feature from the AT model.  
To be consistent with previous works \cite{cheng2019improving,feng2022boosting}, and due to the poor performance, the evaluation of targeted attacks under defense is temporarily omitted.

\begin{table}[tbp]
  \centering
  
  \scalebox{0.75}{
    \begin{tabular}{c|ccccc}
    \hline
    Methods & AT    & BitRed & NRP   & DiffPure & Avg. ASR \Tstrut \\
    \hline
    Signhunter & 36.7  &         38.0  & 28.9 &         12.6  &         26.5  \Tstrut\\
    NES   & 18.6  &         21.0  & 31.3 &         20.0  &         24.1  \Tstrut\\
    $\mathcal{N}$Attack & 39.2  &         45.0  & 34.9 &         24.0  &         34.6 \Tstrut \\
    SimBA & 39.5  &         14.5  & 25.8 &         13.1  &         17.8 \Tstrut \\
    BASES & 0.0     &         \textbf{58.5}  & 28.5  &           8.6  &         31.9 \Tstrut \\
    P-RGF & 43.2  &         38.0  & \underline{37.5}  &         \textbf{30.4}  &         \underline{35.3} \Tstrut \\
    Subspace & \textbf{54.8}  &         \underline{54.0}  & 29.5  &         14.8  &         32.8 \Tstrut \\
    GFCS  & 6.8   &         21.5  & 32.9 &         22.5  &         25.6 \Tstrut \\
    CGA    &   -    &    -   &     -  &  -     & -\Tstrut \\
    \hline
    DifAttack(\textbf{Ours})  & \underline{52.5}  &         46.5  & \textbf{41.5}  &         \underline{28.5}  & \textbf{       42.3 }\Tstrut \\
    \hline
    \end{tabular}%
    }
    \caption{The ASR of AEs from ImageNet against various SOTA defensive techniques in the untargeted setting.}
  \label{tab:ATdefense}%
\end{table}%

\begin{table}[t]
  \centering
  
  \scalebox{0.75}{
    \begin{tabular}{c|ccc}
   
    \hline
     Settings    & $\l_2$    & ASR   & Avg.Q   \Tstrut\\
    \hline
      w/o DF & 57/48/42/48    & 84/96/92/100    & 2,324/1,807/1,779/930 \Tstrut\\
    w/ DF  & \textbf{10/9/9/14}    & \textbf{100/100/100/100}   & \textbf{156/123/179/90}   \Tstrut\\
  
    \hline
    \end{tabular}%
    }
    \caption{The ablation study on the DF module when performing untargeted attacks against ResNet/VGG/GoogleNet/SqueezeNet in ImageNet.}
  \label{tab:withoutDF}%
\end{table}%

\subsection{Ablation Studies}
\subsubsection{DF Module.} To assess the DF module, we train an autoencoder $\mathcal{G}$ w/o DF with the same training mode and architecture as $\mathcal{G}$ w/ DF. Additionally, the disentanglement of adversarial and visual features implemented by DF is realized by randomly dividing the feature $\mathbf{z}$ into two halves. Afterward, we examine the image reconstruction loss and ASR of these two network variants. The experimental results presented in Table \ref{tab:withoutDF} demonstrate that the usage of the DF significantly reduces the $l_2$ reconstruction loss from an average value of 50 to 10, meanwhile improving the attack efficiency. For instance, in the case of ResNet, the ASR increases by 16\%, accompanied by a reduction of over 2,000 queries. 
\subsubsection{Sampling Scale $\tau$.}
Another factor in DifAttack that significantly affects the attack performance is the sampling scale $\tau$ in Eq.(\ref{eq:algo4}). We evaluate the ASR and Avg.Q when changing the value of $\tau$ in targeted attacks, as depicted in Figure \ref{fig:featureSensitivityTargeted}. Generally, as the value of $\tau$ increases, the ASR continuously improves, while the Avg.Q initially decreases and then increases. Based on the trend of Figure \ref{fig:featureSensitivityTargeted}, in targeted attacks, we generally set $\tau$ to the optimal value of most classifiers, which is 12. In untargeted attacks, $\tau$ is usually set to 8, of which the details are provided in the supplementary file.

\subsubsection{Disentangled Features.}
To evaluate the effect of the disentangled adversarial and visual features on the adversarial capability, we randomly sample a noise from a Gaussian distribution $\mathcal{N}(0,\xi^2)$ to perturb either the adversarial or visual feature, while keeping the other feature unchanged. Then we employ our pre-trained autoencoder to decode an image from the fusion of these two features. The resulting ASR of these decoded images is presented in Figure \ref{fig:featureSensitivity}. With the increasing $\xi$, the adversarial feature exhibits a faster growth in ASR compared to the visual feature. This manifestation underscores the high sensitivity of an image's adversarial capabilities to modifications in its adversarial features.

\section{Conclusion}

This work presents a novel score-based black-box attack method named DifAttack, which aims to decouple the adversarial and visual features from an image's latent feature and perform black-box attacks via the disentangled feature space. The main idea is to optimize the adversarial feature while keeping the visual feature unchanged until a successful AE is obtained. Experimental results demonstrate the superior attack efficiency of DifAttack in both close-set and open-set scenarios, as well as against a real-world API.

\section{Acknowledgments}
This work was supported in part by Macau Science and Technology Development Fund under SKLIOTSC-2021-2023, 0072/2020/AMJ and 0022/2022/A1; in part by Research Committee at University of Macau under MYRG2022-00152-FST and  MYRG-GRG2023-00058-FST-UMDF; in part by Natural Science Foundation of China under 61971476; and in part by Alibaba Group through Alibaba Innovative Research Program.

\bibliography{main}

\clearpage
\section{Supplementary Material}

\subsection{Content Outline}
This manuscript contains five sections, offering further elaboration on experimental settings and results that are not presented in the main manuscript due to spatial constraints. The content is organized as follows:
\begin{itemize}
 \item More details about experimental settings.
 	
\item The experimental results and analysis of the targeted attacks in ImageNet and Cifar-10 datasets, with the target class 776 and 4, respectively.
	
\item The experimental results and analysis of the untargeted attacks in the Cifar-10 dataset.	

\item Visualization of Adversarial Examples (AEs) generated by comparative approaches in ImageNet and Cifar-10 datasets.

\item Ablation studies, which delves into the impact of parameter $\tau$ on the attack performance in the untargeted attack setting and explores the usage of different white-box attack methods during the training phase.
 \end{itemize}

\subsection{Experimental Settings}
All of our experiments are conducted using the Nvidia GeForce RTX 3090 graphics card. A single RTX 3090 suffices to accomplish the training process of our designed autoencoder $\mathcal{G}$ in the ImageNet or Cifar-10 dataset with a batch size of 32 or 128, respectively. More details of the training and attack stages are presented in the code.

For performing a black-box attack on Cifar-100 in the open-set scenario, we exclude images from ten categories of Cifar-100 that potentially exhibit semantic overlap with images from Cifar-10. These 
 excluded classes of Cifar-100 encompass vehicles: ``bicycle, bus, lawn mower, motorcycle, pickup truck, rocket, streetcar, tank, tractor, train", which correspond to semantically similar categories in Cifar10, including ``automobile, ship, truck".


 \begin{figure}[t]
\centering
\includegraphics[width=0.45\textwidth]{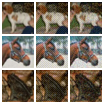} 
\caption{ From left to right, they are clean images, AEs generated by CGA \cite{feng2022boosting}, and AEs crafted by our DifAttack in sequence. These images are randomly sampled from Cifar-10 and produced under untargeted attacks with the $l_\infty$ constraint of $8/255$. CGA introduces a conspicuous grid-like pattern on AEs, while our DifAttack generates adversarial perturbations that align more closely with the inherent distribution of the image.} 
\label{fig:aevisualmini}
\end{figure}

\begin{table*}[htbp]
  \centering
  
  \scalebox{0.75}{
    \begin{tabular}{c|cc|cc|cc|cc|cc|cc|cc|cc}
    \hline
    Datasets & \multicolumn{8}{c|}{ImageNet}                                 & \multicolumn{8}{c}{Cifar-10} \Tstrut\\
    \hline
    Victim Models & \multicolumn{2}{c|}{ResNet} & \multicolumn{2}{c|}{VGG} & \multicolumn{2}{c|}{GoogleNet} & \multicolumn{2}{c|}{SqueezeNet} & \multicolumn{2}{c|}{PyramidNet} & \multicolumn{2}{c|}{DenseNet} & \multicolumn{2}{c|}{VGG} & \multicolumn{2}{c}{ResNet} \Tstrut\\
    \hline
    Methods & ASR   & Avg.Q & ASR   & Avg.Q & ASR   & Avg.Q & ASR   & Avg.Q & ASR   & Avg.Q & ASR   & Avg.Q & ASR   & Avg.Q & ASR   & Avg.Q \Tstrut\\
    \hline
    Signhunter &          58.8  &          7,432  &          77.4  &          6,280  &          33.2  &          8,548  &          67.3  &          6,775  &        \textbf{100.0}  &             567  &        \textbf{100.0}  &           541  &          99.3  &          1,030  &          99.8  &             704  \Tstrut\\
    NES   &          59.8  &          7,731  &          64.3  &          7,301  &          36.2  &          8,636  &          68.8  &          6,956  &          97.3  &          1,218  &          99.1  &           990  &          91.4  &          2,233  &          95.8  &          1,511  \Tstrut\\
    
    $\mathcal{N}$Attck &          99.5  &          4,827  &          97.0  &          4,971  &          70.9  &          6,983  &          96.5  &          4,370  &          99.9  &             778  &        100.0  &           746  &          98.5  &          1,403  &          99.7  &          1,031  \Tstrut\\
   
    SimBA &          97.0  &          3,737  &        \textbf{100.0}  &          \underline{2,898}  &          77.0  &          5,905  &          97.5  &          \underline{3,202}  &          99.2  &             391  &          99.7  &           359  &          96.0  &             601  &          98.6  &             545 \Tstrut \\
    BASES &  32.4    &   7,695    & 23.5     &   8,049   &  10.4     &  9,273     &  21.8     &   8,351     &    85.0   & 1,737       &     88.2   &1,593       &   73.3             & 3,197 &79.6&2,385 \Tstrut\\ 
   P-RGF &        
             59.3  &          5,640  &          74.4  &          4,282  &          43.2  &          6,859  &          71.9  &          4,468    &          98.2  &             423  &          99.4  &           305  &          94.7  &             835  &          98.1  &             387  \Tstrut\\ 
    
    Subspace &          67.3  &          4,701  &          69.9  &          4,392  &          56.3  &          5,789  &          43.2  &          7,146  &          95.1  &          1,373  &          98.4  &           632  &          93.3  &          1,203  &          89.6  &          2,144  \Tstrut\\

    GFCS  &          91.5  &          \underline{3,638}  &          90.3  &          3,266  &          57.5  &          6,749  &          90.0  &          3,590  &          98.4  &             \underline{198}  &          99.5  &           \underline{253}  &          96.6  &            \underline{ 487}  &          97.8  &             \underline{317}  \Tstrut\\

    CGA &          90.5  &          4,803  &          90.8  &          4,998  &          91.3  &          \underline{4,241}  &          92.5  &          3,563  &          98.9  &             798  &             99.8    &   715    &             95.2    & 1,588      &            98.5    &1,034 \Tstrut \\
    \hline
    DifAttack(\textbf{Ours})  &        \textbf{100.0}  &          \textbf{1,912}  &        \textbf{100.0}  &          \textbf{1,566}  &          \textbf{97.5}  &          \textbf{3,310}  &          \textbf{98.5}  &          \textbf{2,052}  &        \textbf{100.0}  &             \textbf{194}  &       \textbf{ 100.0}  &           \textbf{183}  &          \textbf{99.7}  &             \textbf{435}  &        \textbf{100.0}  &             \textbf{223} \Tstrut \\
    \hline
    \end{tabular}%

    }
        \caption{The attack success rate (ASR \%), average number of queries (Avg.Q) of test images in targeted attacks with target class 776 for ImageNet and class 4 for Cifar-10. The best and second best Avg.Q are \textbf{bold} and \underline{underlined}, respectively.}
  \label{tab:targetedOthers}%
\end{table*}%

\subsection{Targeted Attacks with Different Target Labels}\label{app:targeted}
To reduce the randomness of the experimental results in targeted attacks, we select a new target attack class for each dataset that is different from the target class used in Table 1 of the main manuscript. As shown in Table \ref{tab:targetedOthers} below, in the close-set scenarios, we perform targeted attacks with the target class 776 and 4 for ImageNet and Cifar-10, respectively. It can be seen that our DifAttack still largely outperforms other comparative methods. Specifically, in terms of the ASR, our DifAttack achieves nearly perfect (100\%) performance in almost all victim models, which is hard to achieve for other competitors. Meanwhile, the Average Query Number (Avg.Q) also substantially decreases when adopting DifAttack to generate AEs. For example, the Avg.Q obtained by other methods is diminished from 3,638 to 1,912 in our case when attacking ResNet on ImageNet. Note that, the ASR of BASES \cite{cai2022blackbox} in ImageNet remains quite modest. This phenomenon is attributed, as indicated in their paper, to the inherent requirement of a substantial number of surrogate models which is far more than three to attain favorable attack performance. In our experiments, fairly, we only provide three surrogate models for all comparative methods that involve surrogate models.

\begin{table}[t]
  \centering

  \scalebox{0.6}{
    \begin{tabular}{c|cc|cc|cc|cc}
    \hline
    Victim Models & \multicolumn{2}{c|}{PyramidNet} & \multicolumn{2}{c|}{DeseNet} & \multicolumn{2}{c|}{VGG} & \multicolumn{2}{c}{ResNet}  \Tstrut\\
    \hline
    Methods   & ASR   &  Avg.Q  & ASR   & Avg.Q   & ASR   & Avg.Q   & ASR   & Avg.Q  \Tstrut\\
    \hline
    Signhunter & 99.6  & 240   & \textbf{100.0} & 229   & 98.5  & 492   & \textbf{     100.0 } &             187   \Tstrut\\
    NES   & 97.6  & 571   & 97.1  & 696   & 92.5  & 1,251 &          97.9  &             553   \Tstrut\\
    $\mathcal{N}$Attck   & \textbf{100.0} & 354   & \textbf{100.0} & 423   & \textbf{100.0} & 491   & \textbf{     100.0 } &             362 \Tstrut\\
    SimBA   & 98.0  & 327 & 97.8  & 411   & 97.4  & 455 &          98.5  &             285 \Tstrut\\
    BASES   &   98.7    &    195   &   99.4    &   107    &     97.2  &  408     &    98.6   &  183    \Tstrut\\
    P-RGF  & 99.6  & 141   & 99.9  & 134   & 97.1  & 482   &          99.1  &             216  \Tstrut\\
    Subspace   & 99.9  & 183   & \textbf{100.0} & 90    & 98.8  & 247   &          99.7  &             206  \Tstrut\\
    GFCS  & 99.5  & 193   & 99.9  & 154   & 98.1  & 335   &          99.3  &             136  \Tstrut\\
    CGA & \textbf{100.0} & \textbf{34}    & \textbf{100.0} & \textbf{52}    & \textbf{100.0} & \textbf{64}    & \textbf{     100.0 } &              \underline{86}  \Tstrut\\
    \hline
    DifAttack(\textbf{Ours})  & \textbf{100.0} & \underline{61 }   & \textbf{100.0} & \underline{65}    & \textbf{100.0} & \underline{102}   & \textbf{     100.0 } & \textbf{             48 } \Tstrut\\
    \hline
    \end{tabular}%
    }
      \caption{Attack success rate (ASR \%), average number of queries (Avg.Q) of test images from Cifar-10 under the untargeted attack setting with a maximum adversarial perturbation constraint of $\l_\infty=8/255$. The best and second best Avg.Q are \textbf{bold} \underline{underlined}, respectively.}
  \label{tab:untargeted}%
\end{table}%



    

\subsection{Untargeted Attacks on Cifar-10}
We leave the least challenging experiment from the experiments conducted in our work for this section, wherein we perform untargeted attacks on Cifar-10 in the close-set scenario. The experimental results are presented in Table \ref{tab:untargeted}, where our method maintains a 100\% ASR in all cases and achieves comparable query efficiency to the CGA method  \cite{feng2022boosting}. Although our DifAttack has an average query number of 69 on four victim models, slightly higher than the 59 obtained by CGA, DifAttack can still be considered to outperform CGA across the entirety of the black-box attack scenarios according to the following three points: 

1) Regarding the visual effects of adversarial perturbation generated under the same experimental settings in Table \ref{tab:untargeted}, CGA is inferior to our DifAttack in both numerical and visual aspects. Numerically, DifAttack yields adversarial perturbations with an average $l_2$ distance of 1.578, which is lower than the value of 1.731 obtained by CGA; visually, as illustrated in Figure \ref{fig:aevisualmini}, CGA introduces a grid-like pattern on each adversarial image. This regular distortion is easily perceivable to the human eyes, conveying an artificial and abnormal noise. It is also believed that this unnatural noise deviates from the fact that the AEs are designed to be imperceptible to the human eyes. In contrast, our DifAttack generates adversarial perturbations that are more natural and align more closely with the inherent distribution of the image, rendering them inconspicuous to human vision.
2) When it comes to the more challenging targeted attacks on Cifar-10, as can be seen in Table 1 of the main manuscript, the ASR and Avg.Q achieved by our DifAttack are all significantly better than CGA across the four different victim models, where the average query number on four models obtained by DifAttack is only 526, while that of CGA reaches a value up to 1,924. 3) The classification of the small-scale CIFAR-10 dataset and especially the untargeted attack on it are relatively simple tasks. In the large-scale and more practical ImageNet dataset, our attack performance is significantly superior to that of CGA for both untargeted attacks and targeted attacks, particularly in challenging targeted attacks. For instance, in Table \ref{tab:targetedOthers}, when attacking the VGG model in ImageNet, our DifAttack achieves 100\% ASR and 1,566 Avg.Q while CGA only obtains an ASR of 90.8\% and a much higher Avg.Q of nearly 5,000.


\begin{table}[t]
  \centering

  \scalebox{0.6}{
    \begin{tabular}{c|cc|cc|cc|cc}
    \hline
     \multirow{3}{*}{{Settings}} & \multicolumn{4}{c|}{Cifar 10 VGG}          & \multicolumn{4}{c}{ImageNet VGG} \Tstrut\\
\cline{2-9}          & \multicolumn{2}{c|}{Targeted} & \multicolumn{2}{c|}{Untargeted} & \multicolumn{2}{c|}{Targeted} & \multicolumn{2}{c}{Untargeted} \Tstrut\\
\cline{2-9}        & {ASR} & {Avg.Q} & {ASR} & {Avg.Q} & {ASR} & {Avg.Q}  & {ASR} & {Avg.Q}\Tstrut \\
    \hline
    
   $\mathcal{G}$ w/ PGD  & \textbf{100}   & 736   & \textbf{100 }  & \textbf{102}    &    \textbf{100}   &   2,246      &  \textbf{100 }    & \textbf{123}\Tstrut\\ 

   $\mathcal{G}$ w/ MIFGSM &  99.9      &    747     &   \textbf{100 }    & 104 & \textbf{100}&\textbf{2,227}&  \textbf{100}& 147 \Tstrut\\
    
     $\mathcal{G}$ w/ MIFGSM\&PGD &       99.9    &   \textbf{710 } &  \textbf{100 }  &\textbf{102}&\textbf{100}&2,429&  \textbf{100}& 125\Tstrut\\ 
   
    \hline
    \end{tabular}%
    }
      \caption{The ablation study on the white-box attack methods used for generating AEs during training the autoencoder $\mathcal{G}$. This is an example when the victim model is the VGG-16 pre-trained on ImageNet or the VGG-15 pre-trained on Cifar-10 in the close-set attack scenarios. The target classes for ImageNet and Cifar-10 are 864 and 0, respectively.}
  \label{tab:ablationLoss4}%
\end{table}%

 \begin{figure*}[t]
\centering
\includegraphics[width=0.9\textwidth]{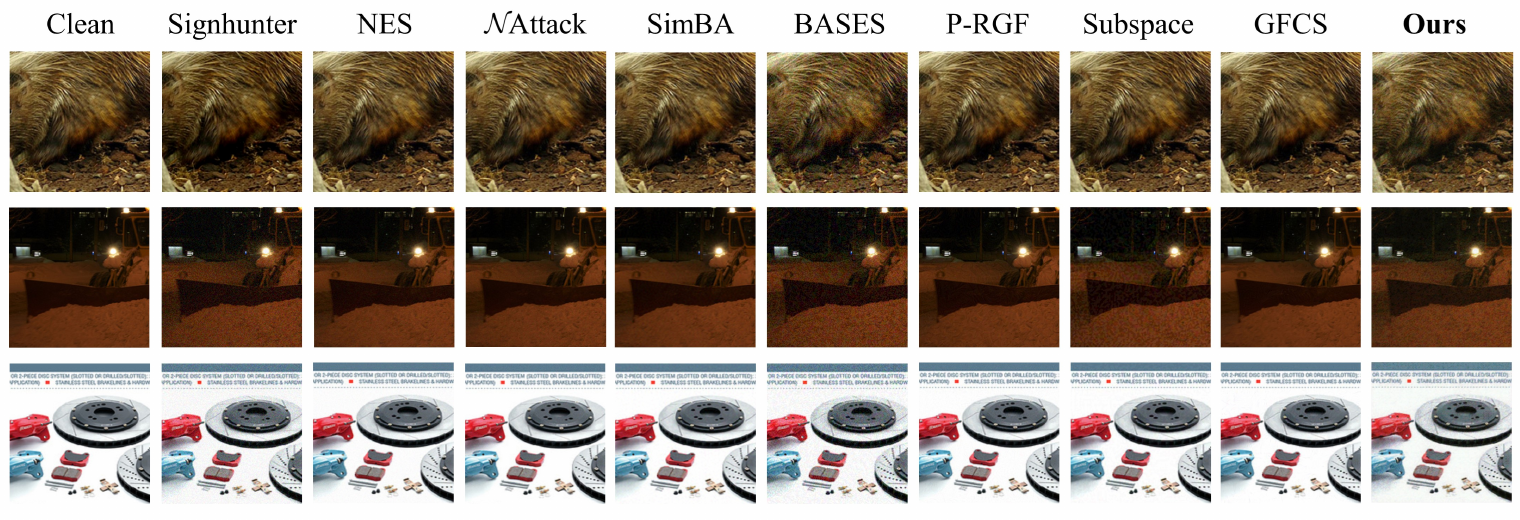} 
\caption{The AEs randomly sampled from ImageNet and generated by different comparative black-box attack methods under untargeted attacks with the $l_\infty$ constraint of $12/255$.}
\label{fig:aevisual}
\end{figure*}

 \begin{figure*}[ht]
\centering
\includegraphics[width=0.9\textwidth]{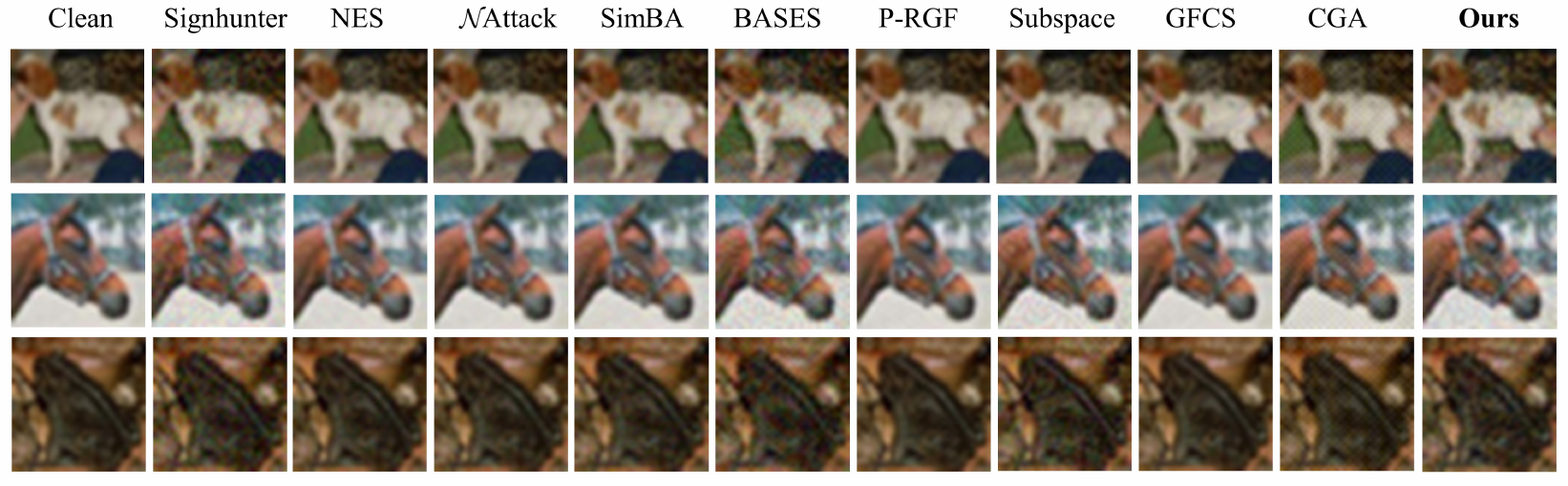} 
\caption{The AEs randomly sampled from Cifar-10 and generated by different comparative black-box attack methods under untargeted attacks with the $l_\infty$ constraint of $8/255$.}
\label{fig:aevisualcifar}
\end{figure*}

 \begin{figure}[t]
\centering
\includegraphics[width=0.45\textwidth]{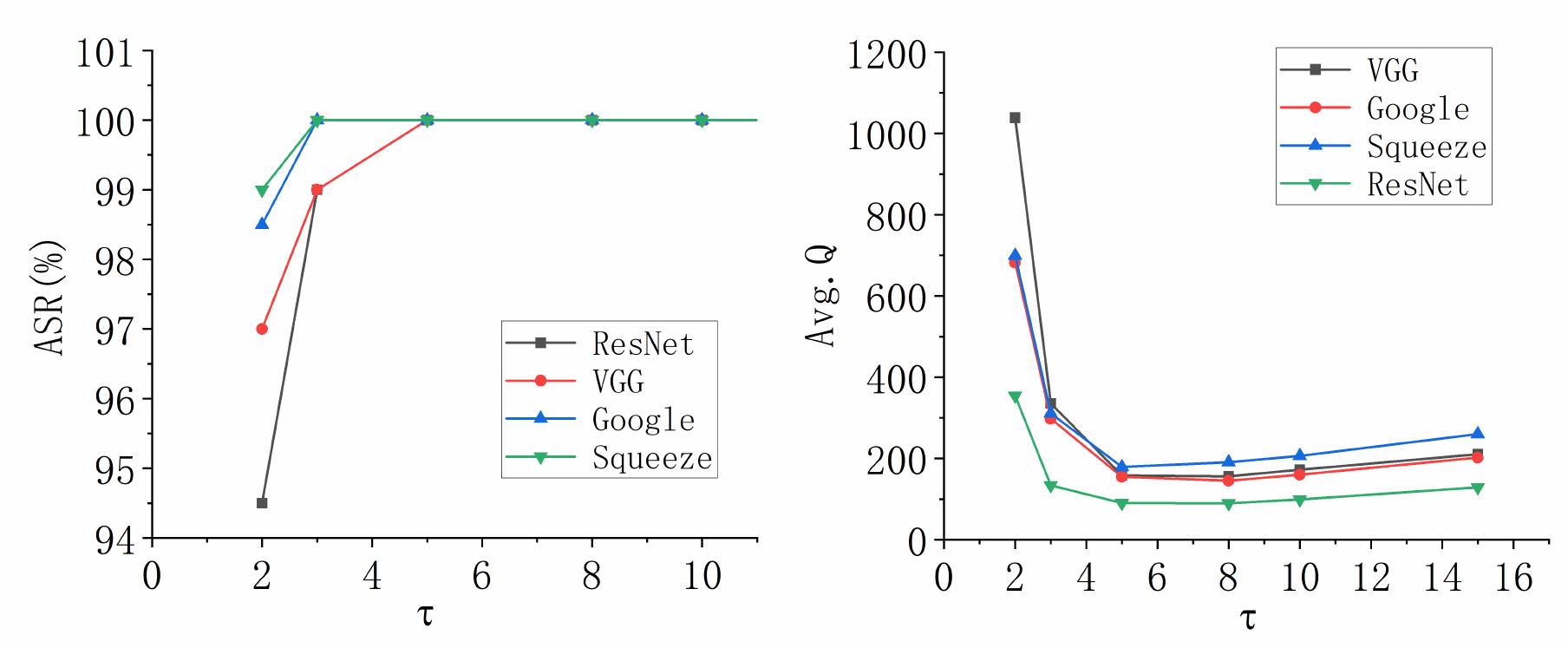} 
\caption{The influence of the parameter $\tau$ on the ASR and Avg.Q when performing untargeted attacks on ImageNet.}
\label{fig:ablationt}
\end{figure}

\subsection{Visualizations of Adversarial Examples}

To compare the generated adversarial perturbation more intuitively, we randomly select three images from the large-scale ImageNet and small-scale Cifar-10 datasets as examples. Then we employ comparative techniques to generate AEs of these images under the constraint of a maximum adversarial perturbation of $l_\infty=12/255$ or $l_\infty=8/255$ for ImageNet and Cifar-10, respectively. The results are shown in Figure \ref{fig:aevisual} for ImageNet and Figure \ref{fig:aevisualcifar} for Cifar-10. It can be observed that the adversarial perturbations generated by our approach exhibit minor visual perception, akin to natural noise in distribution. Notably, since CGA does not provide the training code and their pre-trained models for ImageNet, we are unable to illustrate the AEs of CGA in ImageNet. 

 Overall, considering that our DifAttack is not targeted at generating AEs with minimal perturbation and optimal visual fidelity, but rather aims to reduce query numbers and improve the ASR within certain perturbation constraints. Our approach has succeeded in achieving a commendable balance between visual quality and attack efficiency.
 


\subsection{Ablation Studies}
\subsubsection{The influence of $\tau$ in untargeted attacks.}

A parameter in DifAttack that significantly affects the attack performance is the sampling scale $\tau$ in Eq.(9) of the main manuscript when updating $\mu$. We vary the value of $\tau$ in untargeted attacks while keeping other settings constant. The resulting ASR and Avg.Q are plotted in Figure \ref{fig:ablationt}. It is demonstrated that as the value of $\tau$ increases, the ASR continues to grow until reaching a maximum of 100\%, whereas the Avg.Q initially decreases and then increases. The optimal values of $\tau$ that minimize the query number remain similar for attacking different target models, which range from 5 to 8, and the Avg.Q within this range does not change much. Hence, in practice, we can heuristically set the value of $\tau$ between around 5 and 8 for unknown black-box models. Additionally, combined with the analysis of $\tau$ in the target attack scenario in the main manuscript, we tend to opt for a smaller value of $\tau$ for untargeted attacks, while reserving a larger $\tau$ value for targeted attacks.

\subsubsection{The white-box attack methods in the training stage.}
In the training stage of our DifAttack, we adopt pairs of clean images and their AEs to train an autoencoder for realizing feature disentanglement and image reconstruction. These AEs are generated by performing white-box attacks on one of the collection of surrogate models. Here, we analyze the influence of the chosen white-box attack method for generating such AEs. 

Experimentally, we select existing powerful and efficient white-box attack methods PGD \cite{madry2018towards} and MIFGSM \cite{dong2018boosting} to generate AEs for training. Then PGD and MIFGSM are utilized to construct three different training schemes: the first two involve generating AEs for the entire training set using either PGD or MIFGSM exclusively, denoted as $\mathcal{G}$ w/ PGD and $\mathcal{G}$ w/ MIFGSM respectively, while the third allows us to randomly select between PGD and MIFGSM as the attack method for each batch of images during the training process. Subsequently, we assess the attack performance of DifAttack under these three settings. The experimental results in Table \ref{tab:ablationLoss4} indicate that the choice of attack methods for training has a negligible impact on the ultimate effectiveness of the DifAttack. In other words, our network demonstrates the capacity to effectively disentangle adversarial features and visual features from various types of adversarial perturbation distributions. This also further helps us understand the viewpoint we mention in the main manuscript: the disentanglement of adversarial features is not significantly correlated with the distribution of adversarial examples itself, but rather with the classifiers.

\end{document}